\documentclass[letterpaper, 10 pt, conference]{ieeeconf}  

\IEEEoverridecommandlockouts                              

\usepackage{graphics} 
\usepackage{graphicx}

\usepackage{epsfig} 
\usepackage{amsmath} 
\usepackage{amssymb}  

\usepackage{bm}
\usepackage{mathtools}

\usepackage{algorithm}
\usepackage[noend]{algpseudocode} 
\algnewcommand\AAND{\textbf{ and }}
\algnewcommand\Or{\textbf{ or }}

\usepackage{color}
\usepackage{citesort}
\usepackage{flushend}
\usepackage{url}
\usepackage[breaklinks]{hyperref}
\usepackage[capitalise]{cleveref}
\usepackage{booktabs}
\usepackage{makecell}
\usepackage{multirow}

\usepackage[nolist,nohyperlinks]{acronym}
\acrodef{method}[AOM]{ACRONYM OF METHOD}
\acrodef{gnss}[GNSS]{Global Navigation Satellite System}
\acrodef{ransac}[RANSAC]{Random Sample Consensus}
\acrodef{slam}[SLAM]{Simultaneous Localization And Mapping}
\acrodef{pca}[PCA]{Principal Component Analysis}
\acrodef{ekf}[EKF]{Extended Kalman Filter}
\acrodef{rmse}[RMSE]{Root Mean Square Error} 
\acrodef{ape}[APE]{Absolute Pose Error}
\acrodef{cfar}[CFAR]{Constant False Alarm Rate}
\acrodef{snr}[SNR]{Signal to Noise Ratio}
\acrodef{rcs}[RCS]{Radar Cross Section}
\acrodef{imu}[IMU]{Inertial Measurement Unit}
\acrodef{sgm}[SGM]{Segmi-Global Matching}
\acrodef{dnn}[DNN]{Deep Neural Network}
\acrodef{gru}[GRU]{Gated Recurrent Unit}
\acrodef{hpr}[HPR]{Hidden Point Removal}
\acrodef{raft}[RAFT]{Recurrent All-Pairs Field Transforms}
\acrodef{fov}[FOV]{Field of View}
\acrodef{mclab}[MC-lab]{Marine Cybernetics laboratory}
\acrodef{vio}[VIO]{Visual-Inertial Odometry}
\acrodef{rcm}[RCM]{Refractive Camera Model}

\usepackage{tikz}
\usetikzlibrary{positioning, shapes.geometric, arrows.meta, decorations.pathreplacing, calligraphy}

\DeclareMathAlphabet{\pazocal}{OMS}{zplm}{m}{n}

\newcommand{\Ws}{\pazocal{W}}

\newcommand{\Bs}{\pazocal{B}}

\newcommand{\Vs}{\pazocal{V}}

\DeclareMathAlphabet{\mathpzc}{OT1}{pzc}{m}{it}

\usepackage{array}
\newcolumntype{C}[1]{>{\centering\arraybackslash}p{#1}}
\newcolumntype{M}[1]{>{\raggedright\arraybackslash}p{#1}}

\usepackage{array} 
\newcolumntype{L}[1]{>{\raggedright\let\newline\\\arraybackslash\hspace{0pt}}m{#1}}	
\newcolumntype{S}[1]{>{\centering\let\newline\\\arraybackslash\hspace{0pt}}m{#1}}
\newcolumntype{R}[1]{>{\raggedleft\let\newline\\\arraybackslash\hspace{0pt}}m{#1}}

\makeatletter
\renewcommand*{\@opargbegintheorem}[3]{\trivlist
  \item[\hskip \labelsep{\itshape #1\ #2}] \textit{(#3)}\ }
\makeatother

\title{\LARGE \bf
An Online Self-calibrating Refractive Camera Model with Application to Underwater Odometry
}

\author{Mohit Singh, Mihir Dharmadhikari, and Kostas Alexis 
\thanks{This material was supported by the Research Council of Norway Award NO-327292.}
\thanks{The authors are with the Norwegian University of Science and Technology (NTNU), O. S. Bragstads Plass 2D, 7034, Trondheim, Norway {\tt\small mohit.singh@ntnu.no}}
}

\begin{document}

\maketitle
\thispagestyle{empty}
\pagestyle{empty}

\begin{abstract}
This work presents a camera model for refractive media such as water and its application in underwater visual-inertial odometry. The model is self-calibrating in real-time and is free of known correspondences or calibration targets. It is separable as a distortion model (dependent on refractive index $n$ and radial pixel coordinate) and a virtual pinhole model (as a function of $n$). We derive the self-calibration formulation leveraging epipolar constraints to estimate the refractive index and subsequently correct for distortion. Through experimental studies using an underwater robot integrating cameras and inertial sensing, the model is validated regarding the accurate estimation of the refractive index and its benefits for robust odometry estimation in an extended envelope of conditions. Lastly, we show the transition between media and the estimation of the varying refractive index online, thus allowing computer vision tasks across refractive media.

\end{abstract}

\section{Introduction}\label{sec:intro}
Underwater robots have found application across a growing range of application domains including environmental monitoring~\cite{vasilijevic2017coordinated,schill2018vertex}, search and rescue~\cite{delmerico2019current} and industrial inspection~\cite{shukla2016application}. When operating autonomously, underwater systems tend to employ a diverse set of specialized sensors~\cite{wu2019survey} including ($3\textrm{D}$) sonars, acoustic sensors, doppler velocity log (DVL) devices and \acp{imu}, while vision cameras are typically further assisting the task of the robot without, however, being the prime sensor to support key autonomy tasks such as localization~\cite{bahr2009cooperative,paull2013auv,xu2022robust,johannsson2010imaging}. Although exceptions exist, the above reality is driven by the fact that when navigating through open-ended waters without structures in close proximity, visual data informativeness is bound to degrade significantly. However, the increasing need to inspect underwater structures and other cluttered environments (e.g., submerged oil \& gas facilities, fish farms, kelp forests, underwater caves) stimulates an increasing focus on vision-driven underwater systems~\cite{shukla2016application,ferrera2019aqualoc,miao2021univio,teixeira2020deep,rahman2019svin2,randall2023flsea}. The low-cost of vision-based solutions further strengthens this trend. 

\begin{figure}[ht]
\centering
    \includegraphics[width=0.99\columnwidth]{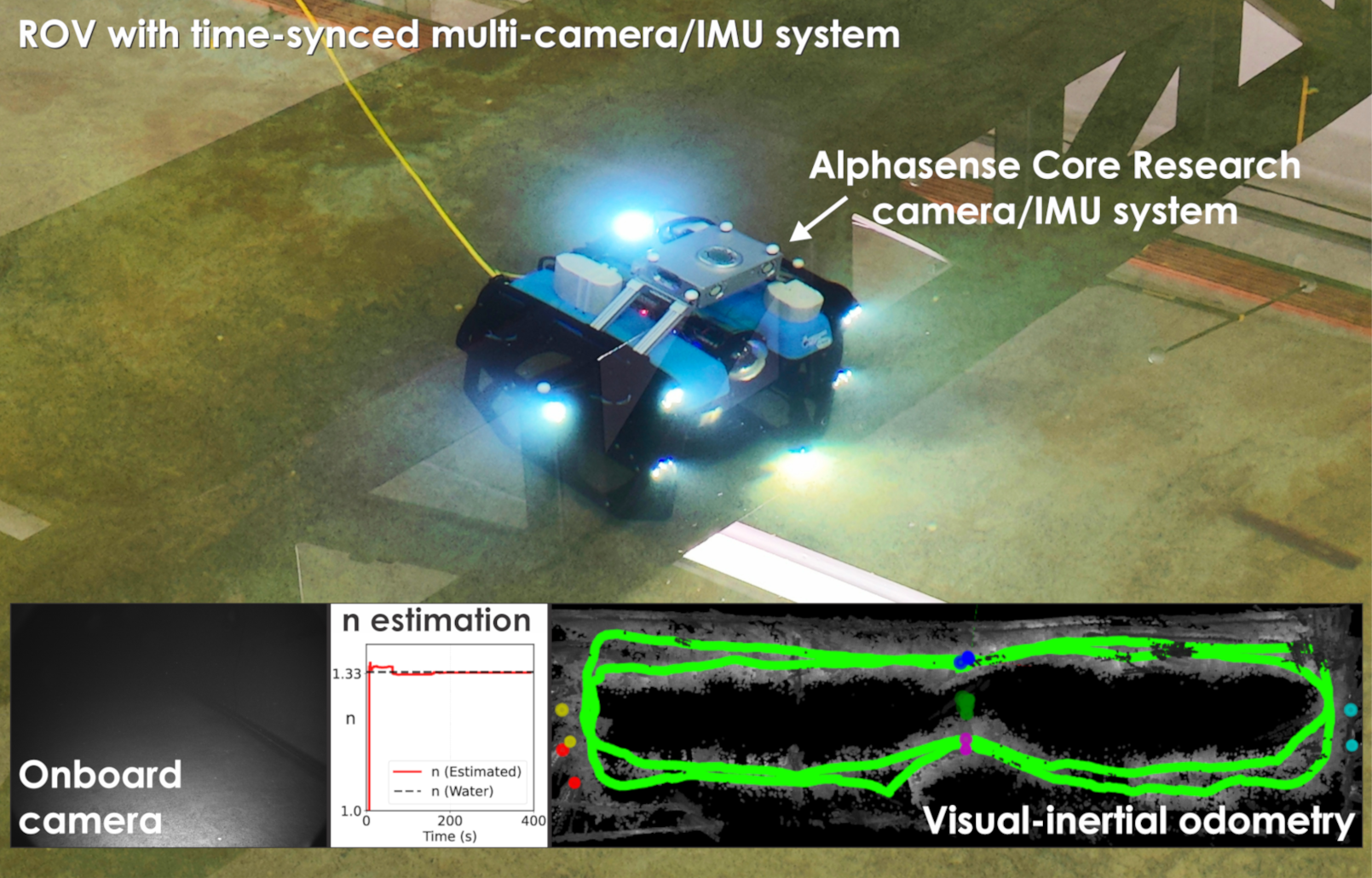}
\vspace{-5ex}
\caption{Instance of one of the experiments involving the proposed self-calibrating refractive camera model, alongside assessing its effect in enabling robust visual-inertial odometry estimation underwater. The method enables the use of cameras and camera-IMU systems in refractive media, while being calibrated conventionally in the air.}
\label{fig:rcm_intro}
\vspace{-5ex}
\end{figure}

Motivated by the above, research has focused on applying visual- and \ac{vio} methods in the underwater domain~\cite{miao2021univio,ferrera2019aqualoc,joshi2023sm}. A common practice among such efforts is the calibration of the camera/\ac{imu} system directly underwater~\cite{randall2023flsea,ferrera2019aqualoc,miao2021univio}, a decision driven by the effect of key phenomena such as the refraction of light in water with the refractive index itself depending on factors such as water salinity, pressure and more. Driven by this observation and aiming to enable the more seamless utilization of underwater vision, this work first contributes a new self-calibrating camera model for refractive media that allows cameras to be calibrated through conventional methods in air and then have an updated camera model that accounts for refraction to be estimated online in the refractive media. The approach not only allows the use of cameras calibrated outside of water, but it also allows adjustment to variations in the refractive index due to physical and environmental conditions during a robot deployment. Second, provided this adaptive camera model, its integration with a state-of-the-art \ac{vio} method allows it to demonstrate superior performance as compared to approaches that employ underwater-calibrated conventional camera models that do not encode the key effects of refractive geometry. Third, the proposed contributions are thoroughly verified in experimental studies where the refractive index is accurately estimated online thus updating the camera model used in \ac{vio} and consequently leading to accurate pose estimation. An extensive dataset is collected to support this study and involves an ROV with a time-synced 5-camera/\ac{imu} set-up swimming in a laboratory pool as in Figure~\ref{fig:rcm_intro}. The data are openly released.

In the remaining paper, Section~\ref{sec:related} presents related work, followed by the proposed camera in Section~\ref{sec:cameramodel} and its integration with underwater visual-inertial odometry in Section~\ref{sec:underwaterperception}. Evaluation studies are detailed in Section~\ref{sec:evaluation}, while conclusions are drawn in Section~\ref{sec:concl}.

\section{Related Work}\label{sec:related}
%
%

This work relates to the domain of adaptive camera modeling for underwater operations, alongside the area of underwater \ac{vio}. Considering the common case that a conventional pinhole camera is observing through a flat window underwater, the work in~\cite{huang2017plate} introduced a plate refractive camera model that includes a pixel-wise variable viewpoint pinhole camera model, the caustic surface for detailing the spatial distribution and viewing directions of various viewpoints, a calibration process that does not require plate removal, and further detailed refraction-based triangulation. The authors in~\cite{treibitz2011flat} focus on non-single viewpoint (non-SVP) camera systems and introduce a physics-based model to improve accuracy. The contribution in~\cite{sedlazeck2012perspective} considers perspective and non-perspective camera models for underwater vision and outlines the limitations of applying the conventional perspective pinhole camera model in such operations. Focusing on estimation, the effort in~\cite{haner2015absolute} addresses the absolute pose estimation problem for a camera observing through a known refractive plane, while highlighting the complexities introduced by Snell's law ambiguities. The works in~\cite{hu2023refractive,hu2021absolute,hu2021absolute} further build formulations for pose estimation in refractive media. Aiming for robust underwater \ac{vio}, the work in~\cite{miao2021univio} utilizes a new underwater image rectification method that separately eliminates water-air refraction distortion and lens distortion using an approximate SVP model. The model is calibrated using underwater camera data, while the presented \ac{vio} method demonstrates good performance. Focusing on self-calibration, the work in~\cite{gu2019environment} offers a camera-\ac{imu} calibration model allowing intrinsic and extrinsic parameters of a monocular set-up to be estimated inside the water with the camera facing a calibration target. Focusing on fiducial-based localization underwater, the authors in~\cite{zhang2021open} explicitly consider a refractive camera model to improve accuracy. Most commonly, state-of-the-art works in underwater \ac{vio} employ the practice of calibrating their sensor set-ups underwater, either on shallow water and then used elsewhere or even directly in the area of interest~\cite{shkurti2011state,joshi2023sm,ferrera2019aqualoc,randall2023flsea,miao2021univio}. Retaining vision as the sole exteroceptive modality, the authors in~\cite{hu2022tightly} further fuse pressure data in a tightly-coupled manner thus assisting \ac{vio} consistency. Compared to this body of works, this paper first contributes a new self-calibrating camera model that allows to depart from the conventional resource-heavy process of calibrating a camera (included in its waterproof casing and typically flat protective window) underwater in the area of a particular deployment (as the refractive index depends on factors such as salinity and pressure) and instead seamlessly calibrate a stereo camera pair above water (outside of its protective case) and then estimate online --during deployment-- the refractive index of the medium to self-adjust the camera model appropriately. This contribution is then combined with further work on tailored underwater \ac{vio} and depth estimation.

\section{Self-calibrating Refractive Camera Model}\label{sec:cameramodel}
\begin{figure}[ht]
\centering\includegraphics[width=0.70\columnwidth]{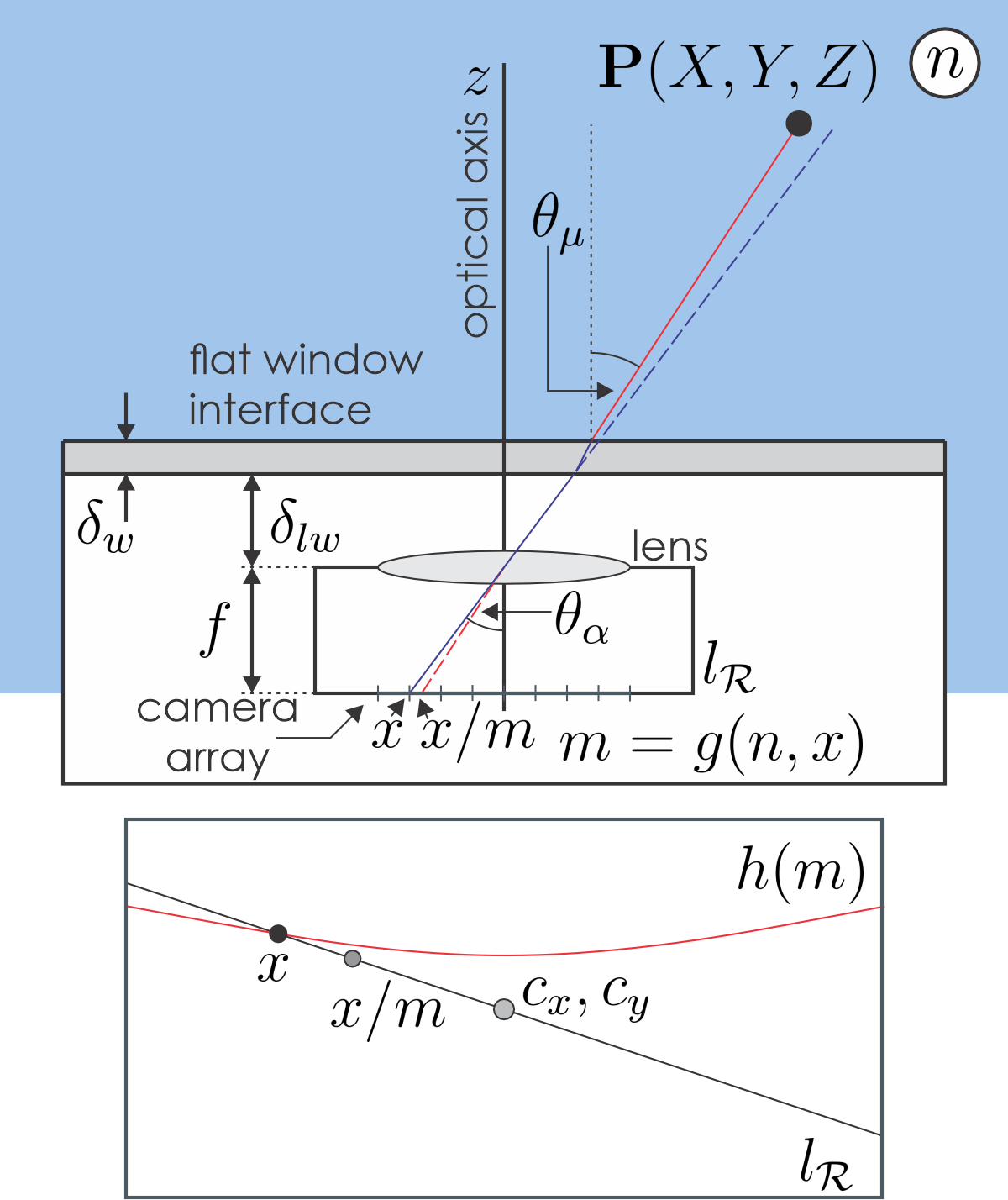}
\vspace{-1ex}
\caption{Visualization of how a ray from an object at a 3D location in the refractive medium intersects the flat interface and proceeds through the air chamber to reach the camera lens and be captured by the sensor. }
\label{fig:camera_geom_2d}
\vspace{-2ex}
\end{figure}

The proposed \ac{rcm} is tailored to vision-based robot operations in refractive media such as water as visualized in Figure~\ref{fig:camera_geom_2d}. The model applies to stereo vision systems using thin and flat transparent plate windows as part of the proofing case of the cameras with small distances between the camera lenses and the refractive interface. These assumptions can be ensured by sensor construction.  

The contribution involves two components. First, the proposed refractive camera model is formulated by incorporating the refractive index $n$ of a medium as a parameter. Second, given the \ac{rcm} we outline a method that allows the online estimation of $n$ and thus the self-calibration and adaptation of the model through feature association on stereo pairs. 

We assume that the pinhole camera model and the refractive interface have negligible separation ($\delta_w$ and $\delta_{lw}$ shown in Figure~\ref{fig:camera_geom_2d} are small). This allows us to relate the incident ray (from a 3D point $\textbf{P}(X,Y,Z)$) in a refractive medium with index $n$ which relates $\theta_{\mu}$ from the optical axis, and the corresponding refracted ray with $\theta_{\alpha}$ from the optical axis in the camera enclosure using Snell's law as following:

\begin{equation}
    \sin{\theta_{\alpha}}=n\sin{\theta_{\mu}}
\label{eq:snell's law}
\end{equation}
Given this, an observed point $x$ on camera coordinates can be related to an ideal undistorted point $x/m$ where $m=g(n, x)$ a function that accounts for the distortion due to refraction as the radial distortion factor. Further, for a given focal length $f$ of a pinhole camera it holds:

\begin{equation}
    x/f = \tan{\theta_{\alpha}},~x/(mf)=\tan{\theta_{\mu}}
\label{eq:normalized_coords_tan}
\end{equation}
Therefore, it holds that
\begin{equation}
    m = \tan{\theta_{\alpha}}/\tan{\theta_{\mu}}
\label{eq:m_to_tan}
\end{equation}

Let us re-write Eq.~(\ref{eq:m_to_tan}) using Eq.~(\ref{eq:snell's law}):
\begin{equation}
    m = \sqrt{\frac{n^2 - \sin^{2}{\theta_{\alpha}}}{1 - \sin^{2}{\theta_{\alpha}}}}
\end{equation}

Given, $\sin(\theta_{\alpha})=\sqrt{\frac{\bar{x}^2}{1+\bar{x}^2}}$ where $\bar{x}=x/f$. Hence 

\begin{equation}\label{eq:mform1d}
    m = \sqrt{n^2\bar{x}^2 + n^2- \bar{x}^2}
\end{equation} 
for a 1D camera on the plane $l_\mathcal{R}$. We extend this result to a 2D camera in the following subsection.

\subsection{Refractive Camera Model}

Let $\textbf{K}$ be an ideal camera matrix for a pinhole model:

\begin{equation}
\textbf{K}=\left[\begin{array}{ccc}f_x & 0 & c_x \\ 0 & f_y & c_y \\ 0 & 0 & 1\end{array}\right]
\end{equation}
where $f_x$ and $f_y$ are the focal length and $c_x$, $c_y$ are the image center coordinates. Without loss of generality, we assume the pixel coordinate frame is shifted to the origin, hence $c_x = 0$, $c_y = 0$. For the following we assume $3$ coordinate frames for the camera model, $\textbf{p}=[p, q, 1]^{\top}$ an ideal pinhole camera model for air, $\textbf{u}=[u, v, 1]^{\top}$ an ideal pinhole camera model for the refractive media, 
$\textbf{x}=[x, y, 1]^{\top}$ the observed pixel coordinates. We use the $(\bar{.})$ notation for normalized variables, where capital letters are normalized by Z in 3D coordinates, while small letter variables are normalized by focal length $f$ in image coordinates. Considering a stereo pair, we assume that the camera parameters, namely intrinsics and extrinsics are known and the images are rectified in for lens distortion. Thus, in an ideal case of no refraction, it holds $\textbf{p} = \textbf{u} = \textbf{x}$. Then using Eq.~(\ref{eq:normalized_coords_tan}),(\ref{eq:m_to_tan}) the camera projection can be written for a 3D point in the camera frame $\textbf{P}(X, Y, Z)$ as:

\begin{equation}
    \left[\begin{array}{l}x \\ y \\ 1\end{array}\right]=\left[\begin{array}{ccc}m f_x & 0 & 0 \\ 0 & m f_y & 0 \\ 0 & 0 & 1\end{array}\right]\left[\begin{array}{c}X / Z \\ Y / Z \\ 1\end{array}\right]
\end{equation}

Note that although this equation resembles the general pinhole model projection, it is a nonlinear mapping due to $m$. Thus, we must decouple a linear pinhole model for refractive media and corresponding distortion model. We rewrite:

\begin{equation}
    \left[\begin{array}{l}x \\ y \\ 1\end{array}\right]=\left[\begin{array}{lll}1 & 0 & 0 \\ 0 & 1 & 0 \\ 0 & 0 & \frac{n}{m}\end{array}\right]\left[\begin{array}{ccc}n f_x & 0 & 0 \\ 0 & n f_y & 0 \\ 0 & 0 & 1\end{array}\right]\left[\begin{array}{l}\bar{X} \\ \bar{Y} \\ 1\end{array}\right]
\end{equation}
where the last two matrices represent an ideal pinhole model in the refractive medium with index $n$. Therefore:

\begin{equation}
    \left[\begin{array}{l}u \\ v \\ 1\end{array}\right]=\left[\begin{array}{ccc}n f_x & 0 & 0 \\ 0 & n f_y & 0 \\ 0 & 0 & 1\end{array}\right]\left[\begin{array}{l}\bar{X} \\ \bar{Y} \\ 1\end{array}\right]
\end{equation}

Using the above equations we can write:
\begin{equation}
    \left[\begin{array}{l}x \\ y \\ 1\end{array}\right]=\left[\begin{array}{ccc}1 & 0 & 0 \\ 0 & 1 & 0 \\ 0 & 0 & \frac{n}{m}\end{array}\right]\left[\begin{array}{l}u \\ v \\ 1\end{array}\right]=\left[\begin{array}{c}u \\ v \\ \frac{n}{m}\end{array}\right]
\label{eq:cam_model_equivalence}
\end{equation}

Lastly, rearranging gives: $[u, v, 1]^{\top}=[x, y, \mathrm{~m_n} ]^{\top}$ where $\mathrm{m_n} = m/n$. Note that this equivalence holds in the homogeneous coordinates. Therefore, for a given refractive index $n$ the image can be un-distorted for computer vision tasks. The following subsections describe the estimation for refractive index for a stereo camera setup. 

\subsection{Derivation of Refractive Index for Stereo Camera Pair}
Let $\textbf{K}_{l} = \textrm{diag}(n f_x, n f_y, 1)$ and $\textbf{K}_{r} = \textrm{diag}(n f_x, n f_y, 1)$ be the new pinhole model for a left camera and right camera of a stereo pair, respectively, and $\textbf{R}, \textbf{t}$ is the known relative pose among the two cameras. Then the fundamental matrix for the new system can be written as:
\begin{equation}
\mathbf{F}=\mathbf{K}_r^{-T}[\mathbf{t}]_{\times} \mathbf{R} \mathbf{K}_l^{-1}
\label{eq:epipolar_constraint}
\end{equation}
Hence, the epipolar constraint can be written as $\mathbf{u}_r^T\mathbf{F}\mathbf{u}_l = 0$ (where subscripts $r,l$ represent right and left camera respectively). Using Eq.~(\ref{eq:cam_model_equivalence}) we can write $\left[x_r, y_r, m_{n, r}\right] \mathbf{F}\left[x_l, y_l, m_{n, l}\right]^{T}=0$.

For an ideal stereo pair the translation is $\mathbf{t} = [t_x, 0, 0]^{T}$ and the rotation $\mathbf{R} = \mathbf{I}_{3\times3}$, Thus, without loss of generality, Eq.~(\ref{eq:epipolar_constraint}) can be simplified to obtain:

\small
\begin{equation}
\left[x_{r}, y_{r}, m_{n,r}\right] \left[\begin{array}{lll}0 & 0 & 0 \\ 0 & 0 & \frac{-t_x}{nf_y}\\ 0 & \frac{t_x}{nf_x} & 0\end{array}\right]\left[x_l, y_l, m_{n,l}\right]^{T}=0
\end{equation}
\normalsize
Therefore: 

\small
\begin{equation}
    \frac{y_l m_{n,r}}{f_{y,l}}=\frac{y_{r}m_{n,l}}{f_{y,r}}
\label{eq:epipolar_line_equated}
\end{equation}
\normalsize
Substituting $m = \sqrt{n^2\bar{r}^2 + n^2- \bar{r}^2}$ (Eq.~(\ref{eq:mform1d} but in 2D) where $\bar{r}=\sqrt{\bar{x}^2+\bar{y}^2}$ we get:

\small
\begin{equation}
    n = \sqrt{\frac{\bar{y}_l^2\bar{r}_{r}^{2} - \bar{r}_l^{2}\bar{y}_{r}^{2}}{\bar{y}_{l}^2\bar{r}_{r}^{2} + \bar{y}_{l}^2 - \bar{r}_{l}^2\bar{y}_{r}^{2} - \bar{y}_r^2}}
\label{eq:n_stereo}
\end{equation}
\normalsize
which in turn gives rise to the online estimation of the refractive index $n$ using a stereo camera. 

\subsection{Locus of a Point Correspondence}

Given an observed point $(x_l, y_l)$ in the left image the locus of the corresponding point $(x_r, y_r)$ can be written as following by re-arranging Eq.~ (\ref{eq:epipolar_line_equated}):

\small
\begin{equation}
    \frac{y_l}{m_{n,l}f_l}=\frac{y_r}{m_{n,r}f_r}=S=\bar{v}_l=\bar{v}_r
\end{equation}
\normalsize
where $S$ is a constant. We further expand the terms using Eq.~(\ref{eq:mform1d}) and rearrange to obtain the locus of $(x_r, y_r)$ as:

\small
\vspace{-2ex}
\begin{equation}\label{eq:hyperbola}
    h(m)=(n^2-1)\bar{x}_{r}^2 + (n^2-1-\frac{1}{S^2})\bar{y}_{r}^2 + n^2=0
\end{equation}
\normalsize
which is the equation of a hyperbola, essentially, this is the epipolar line warpped by the refractive distortion (Figure~\ref{fig:camera_geom_2d}).

\subsection{Refractive Index Observability} \label{subsec:observability}


The estimation of $n$ from Eq.~\ref{eq:n_stereo} using $(x_l, y_l), (x_r, y_r)$  is affected by the uncertainty in correspondence. It will be perfect when the matched $(x_r,y_r)$ lies on $h(m)$ corresponding to the true refractive index $n_0$. If the estimated $n$ is to lie in $n_0 \pm \delta_n$ ($\delta_n$ a small finite deviation), the matched $(x_r, y_r)$ must lie within the two $h(m)$ passing through $(x_l, y_l)$, $[h(m)_{-\delta_n}, h(m)_{\delta_n}]$, and corresponding to the refractive indices ($n_0 \pm \delta_n$). The smaller the angle $\beta$ between the tangents to the $[h(m)_{-\delta_n}, h(m)_{\delta_n}]$ at $(x_l, y_l)$, the smaller the separation between them. This makes the $n$ estimation more susceptible to uncertainty in correspondences. For points close to the $x$ axis, the two hyperbolae are flat and for those close to the $y$ axis, they are tangential to each other resulting in large errors $e_n = |n-n_0|$ in estimated $n$. The trend of $\beta$ and $e_n$ can be seen in Figure~\ref{fig:method_outline}.i and ii respectively. This analysis motivates for a mask to exclude the refractive index estimation form pixels near $x$ and $y$ axis.





\subsection{Online Refractive Index Estimation}
\begin{figure}[ht]
\centering\includegraphics[width=0.99\columnwidth]{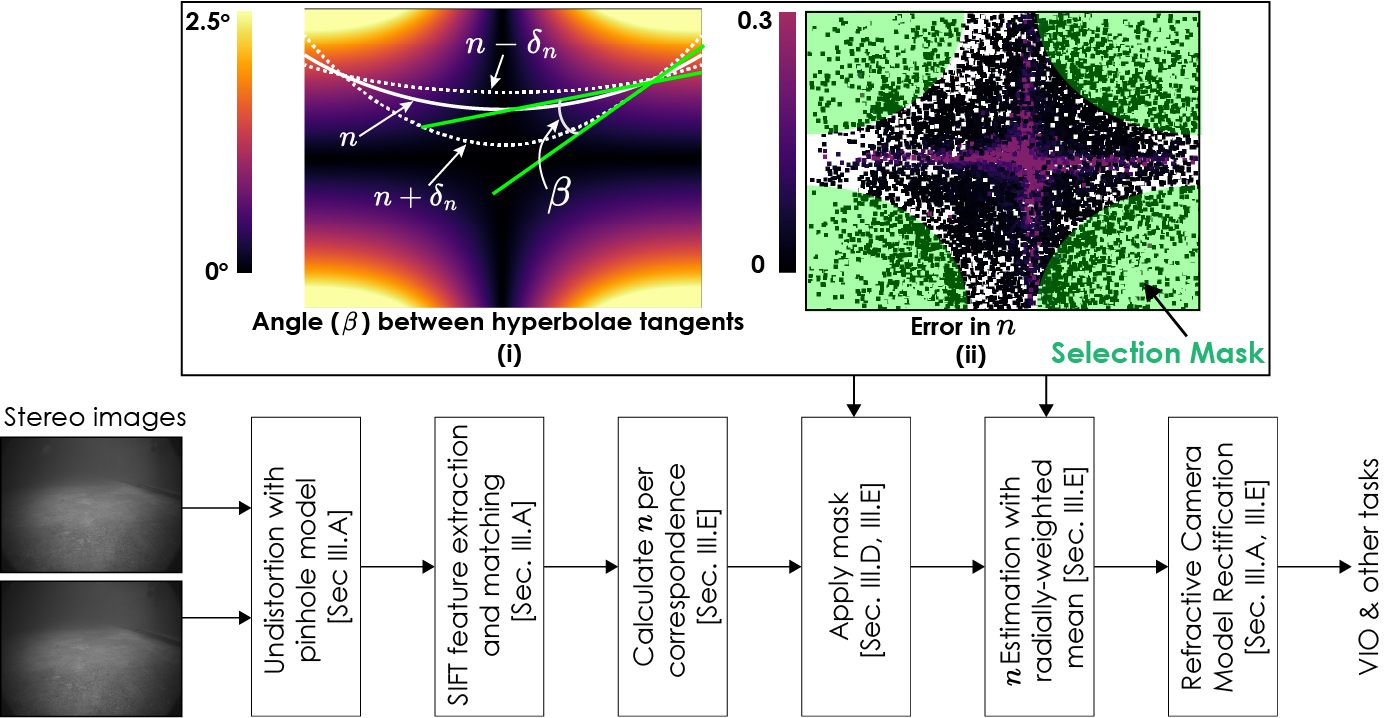}
\vspace{-5ex}
\caption{Signal flow of the proposed approach for self-calibrating refractive camera model estimation. The plot for the angle between the tangents of the intersecting hyperbolae representing the observability of refractive index and the error plot over the image allow to improve the accuracy of online estimation (obtained from data) discussed in Sub-Section(\ref{subsec:observability})}.
\label{fig:method_outline}
\vspace{-4ex}
\end{figure}

Given the above background, we now detail the pipeline of online refractive index estimation. The overall pipeline of the proposed approach for self-calibrating refractive camera model derivation is outlined in Figure~\ref{fig:method_outline}. Given the stereo image pair at a time instant $t$, the method first rectifies them for lens distortion and orientation offset. Note that due to the low-light conditions often encountered underwater the images are first processed through Contrast Limited Adaptive Histogram Equalization (CLAHE)~\cite{zuiderveld1994contrast}. Next, we extract and match SIFT~\cite{lowe1999object} feature correspondences $\mathbb{S}$ between the stereo pair. Motivated by the observability analysis in Section~\ref{subsec:observability}, we employ a masking function to remove feature from areas having low observability of $n$ to generate the filtered set $\mathbb{S}_f\subset \mathbb{S}$. In our implementation, the mask $\mathcal{M}$ takes the form $\sqrt{|u|}\sqrt{|v|}>\eta_{thr}$ (where $0<\eta_{thr}<1$ is a constant which was set to $0.3$ in the experiments). 

Next, $n$ is estimated for all feature correspondences in $\mathbb{S}_f$ using Eq.~(\ref{eq:n_stereo}) and added to a buffer $\mathbb{B}$ containing $n$ values estimated from all selected features upto time $t$. The final output $n$ is calculated through a radially-weighted mean with the weights $r_i$ being the distance of the feature from the optical axis in the left image plane as: 

\small
\vspace{-1ex}
\begin{equation}
    n = \frac{\sum{r_i n_i }}{\sum{r_i }}, \forall n_i \in \mathbb{B}
\end{equation}
\normalsize

This estimated $n$ is then used to rectify the images online for \ac{vio} and can also be used for further vision tasks.

\section{Underwater Visual-Inertial Odometry}\label{sec:underwaterperception}
The proposed refractive camera model was combined with a state-of-the-art visual-inertial odometry method thus leading to a solution tailored to resilient underwater localization. Specifically, ROVIO~\cite{bloesch2015robust} was selected to be used motivated by its good low-light performance as evaluated in~\cite{cerberus_science,cerberus_finals,cerberus_tunnel_urbam}. 

ROVIO combines multi-level image patch tracking with an Extended Kalman Filter (EKF) and employs QR-decomposition to reduce error dimensionality, ensuring computational efficiency in the Kalman filter update. Although originally ROVIO uses direct image intensity errors to derive the filter innovation term, in this work the traditional formulation of the innovation term including reprojection calculations was utilized. ROVIO's approach is robocentric, estimating landmarks relative to the camera pose. This study utilizes a stereo system and leverages cross-correlation between keypoints in both cameras. Estimated landmarks are decomposed into a bearing vector and a depth parametrization using an inverse depth formula. The method considers the \ac{imu}-fixed coordinate frame $\Bs$, the camera-fixed frame $\Vs$, and the inertial frame $\Ws$, resulting in a state vector $\mathbf{s}$ of dimension $l$ and associated covariance $\Sigma$:

\small
\begin{equation}
    \mathbf{s} = \left [ \mathbf{r}~\mathbf{q}~\mathbf{v}~\mathbf{b}_f~\mathbf{b}_\omega~\mathbf{c}_L~\mathbf{c}_R~\mathbf{z}_L~\mathbf{z}_R | \boldsymbol{\mu}_0,....,\boldsymbol{\mu}_J~\rho_0,...,\rho_J \right]
\end{equation}
\normalsize
where $\mathbf{r},\mathbf{v}$ are the robocentric position and velocity of the \ac{imu} expressed in $\Bs$, $\mathbf{q}$ is the \ac{imu} attitude represented as a map from $\Bs\rightarrow \Ws$, $\mathbf{b}_f,\mathbf{b}_\omega$ are the accelerometer and gyroscope bias expressed in $\Bs$, $\mathbf{c}_L,\mathbf{c}_R,\mathbf{z}_L,\mathbf{z}_R$ are the translational and rotational components of the left and right camera extrinsics against the \ac{imu} represented as maps from $\Bs\rightarrow \Vs$, $\boldsymbol{\mu}_j$ is the bearing vector to the $j$-th feature expressed in $\Vs$ and $\rho_j$ is the associated depth parameter such that the feature distance $d_j$ takes the form $d(\rho_j) = 1/\rho_j$. Note that in this work, the extrinsics $\mathbf{c}_L,\mathbf{c}_R,\mathbf{z}_L,\mathbf{z}_R$ are identified offline and set to these fixed values throughout the tests. As necessary implementation difference, ROVIO is interfaced with the undistorted images from the \ac{rcm} which is updated online as described in Section~\ref{sec:cameramodel}. 



\section{Evaluation Studies}\label{sec:evaluation}

A set of experimental studies were conducted to evaluate the proposed \ac{rcm} and the overall approach. 

\subsection{Robot Experimental Set-up}

To verify the proposed contributions, an underwater robot was developed integrating a tightly-synchronized multi-camera/\ac{imu} system. Specifically, the BlueROV platform was employed onboard which a) an Alphasense Core Research Development Kit, and b) an NVIDIA Orin compute board were integrated. Alphasense integrates on a rigid frame with five monochrome Sony IMX-287 global-shutter cameras with 0.4MP resolution. Each camera has a \ac{fov} setting with opening angle $D\times H\times V = 165.4^\circ \times 126^\circ \times 92.4^\circ$ and a focal length of $2.4\textrm{mm}$. The cameras are tightly-synchronized with a Bosch BMI085 \ac{imu} using a mid-frame, exposure-compensated approach ensuring synchronization accuracy no worse than $100\mu \textrm{s}$. The system further offers time sync with the NVIDIA Orin via PTP. Data is transmitted through a Gigabit Ethernet interface. Both Alphasense and the NVIDIA board are separately waterproofed through specialized casing and mounted onboard the BlueROV. The Alphasense is mounted on the robot's top with a down pitch inclination angle of $16^\circ$. The robot is shown in Figure~\ref{fig:rcm_intro}. 


\begin{figure*}[ht!]
\centering
    \includegraphics[width=0.99\textwidth]{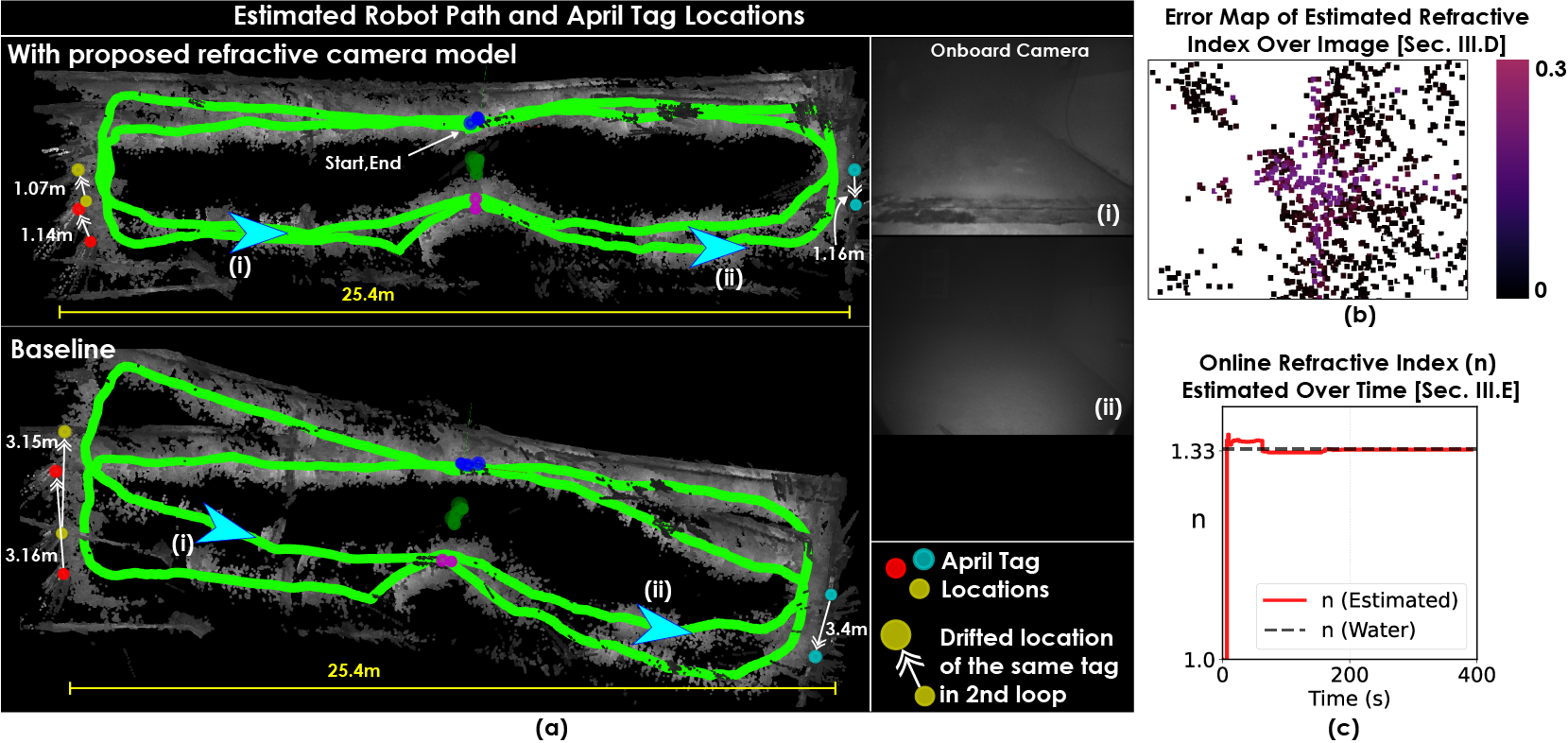}
\vspace{-2ex} 
\caption{Detailed results of one of the trajectories in Trajectory Group $1$. The ambient lighting was set to the lowest level whereas the onboard lighting was at the highest level. Sub-figure (a) shows a qualitative comparison between the odometry solution estimated using the proposed \ac{rcm} and the baseline ROVIO solution. The colored circles show the locations of the detected AprilTag sets. The locations of the tag-sets are shown each time that set is detected and the error in these locations is used to measure the quality of the odometry. This and the visible skewing of the path clearly show that odometry estimated using the proposed model is more accurate than the common approach of calibrating an otherwise conventional camera model directly underwater (which is also more laborious). Sub-figure (b) shows pixels in the input image the features from which were used to calculate the refractive index ($n$) colorized based on the error in the calculated $n$ versus the true value for water ($1.33$). Finally, sub-figure (c) shows the estimated $n$ over the duration of the mission.}
\label{fig:detailedresult}
\vspace{-2ex} 
\end{figure*}

\subsection{The MC-lab Underwater Dataset}

Using the described robotic system, a set of experiments were conducted in the \ac{mclab} of NTNU. \ac{mclab} offers a water tank with dimensions $L \times B \times D = 40 \textrm{m} \times 6.45\textrm{m} \times 1.5 \textrm{m}$. A set of experiments were conducted with the robot performing different trajectories presenting varying difficulty and recording stereo camera frames (from the front-facing cameras) and \ac{imu}. 

In further detail, a total of $24$ missions were collected. These missions were classified into $4$ groups with each group having a different motion pattern as shown in Figure~\ref{fig:collective_figure}. All missions begin with the robot being in the center of the tank (along the length) and consist of two laps of approximately the same motion pattern. In the trajectories in the first group, the robot is piloted along a rectangular shape along the walls of the tank with the camera pointing along the path. Some visual features on the walls are always visible to the camera. In the second group, the robot is moved in a figure-$8$ fashion with the camera pointing along the path. As the robot passes through the center point of the figure-$8$, the number of visual features seen by the camera reduces significantly. The third group consists of trajectories in which the robot predominantly stays in the center of the tank (width-wise) as it moves along the length of the tank. The camera is pointed along the path and only sees features on the ground for the majority of the trajectory. Finally, in the last group, the robot is moved along a similar path as in Group $3$ but the camera orientation is perpendicular to the motion.
Three different onboard illumination levels combined with two ambient lighting levels were tested resulting in each trajectory in the group having a different lighting condition.

In a few selected spots, sets of AprilTags were added such that partial ``groundtruth'' can be acquired to access the robot's motion. These tag-sets are placed such that the robot is not observing them in the majority of any of its missions. A total of $6$ sets were placed, three were placed at the center of the tank facing downwards and the other three were distributed at the two ends of the tank. Figure~\ref{fig:detailedresult} shows the location of the tag-sets as reported by the onboard odometry solution.
As each tag-set is seen at least twice, the difference in the position between the two (or more) detections is used as a metric to evaluate the performance of the odometry solutions.

To enable research reproducibility and verifiability, the derived dataset is openly released at \url{https://github.com/ntnu-arl/underwater-datasets}.

\subsection{Detailed Trajectory Evaluation}

A trajectory belonging to the first group among the abovementioned categories is considered to detail the results of estimating the refractive camera model and its beneficial effects for robust odometry estimation. Figure~\ref{fig:detailedresult} presents the estimation of the refractive index as the robot navigates in the MC-lab, the odometry results of ROVIO when equipped with the image corrected by \ac{rcm} and the ``Baseline'' results of ROVIO with a conventional pinhole equidistant model calibrated underwater. The extrinsics of camera-IMU are identical between the two results and so holds for all other ROVIO parameters. As shown in Figure~\ref{fig:detailedresult}, the proposed approach rapidly converges to an accurate estimate of the refractive index of water, estimated at $n=1.332082$ (nominal value $n=1.33$) which in turn allows for accurate estimation of odometry. This is significantly better than the baseline ROVIO result when the conventional camera model and calibration approach are employed. The accuracy can be verified both a) qualitatively by observing the shape of the path (and the developing drift in the baseline ROVIO solution) as well as the point cloud of the MC-lab pool, reconstructed using RAFT-Stereo~\cite{lipson2021raftstereo} for disparity calculation and the odometry estimates of our solution, and b) quantitatively based on the odometry errors developed when the deployed AprilTags are re-observed in the mission. 

\subsection{Collective Validation Results}

\begin{figure}[ht!]
\centering
    \includegraphics[width=0.99\columnwidth]{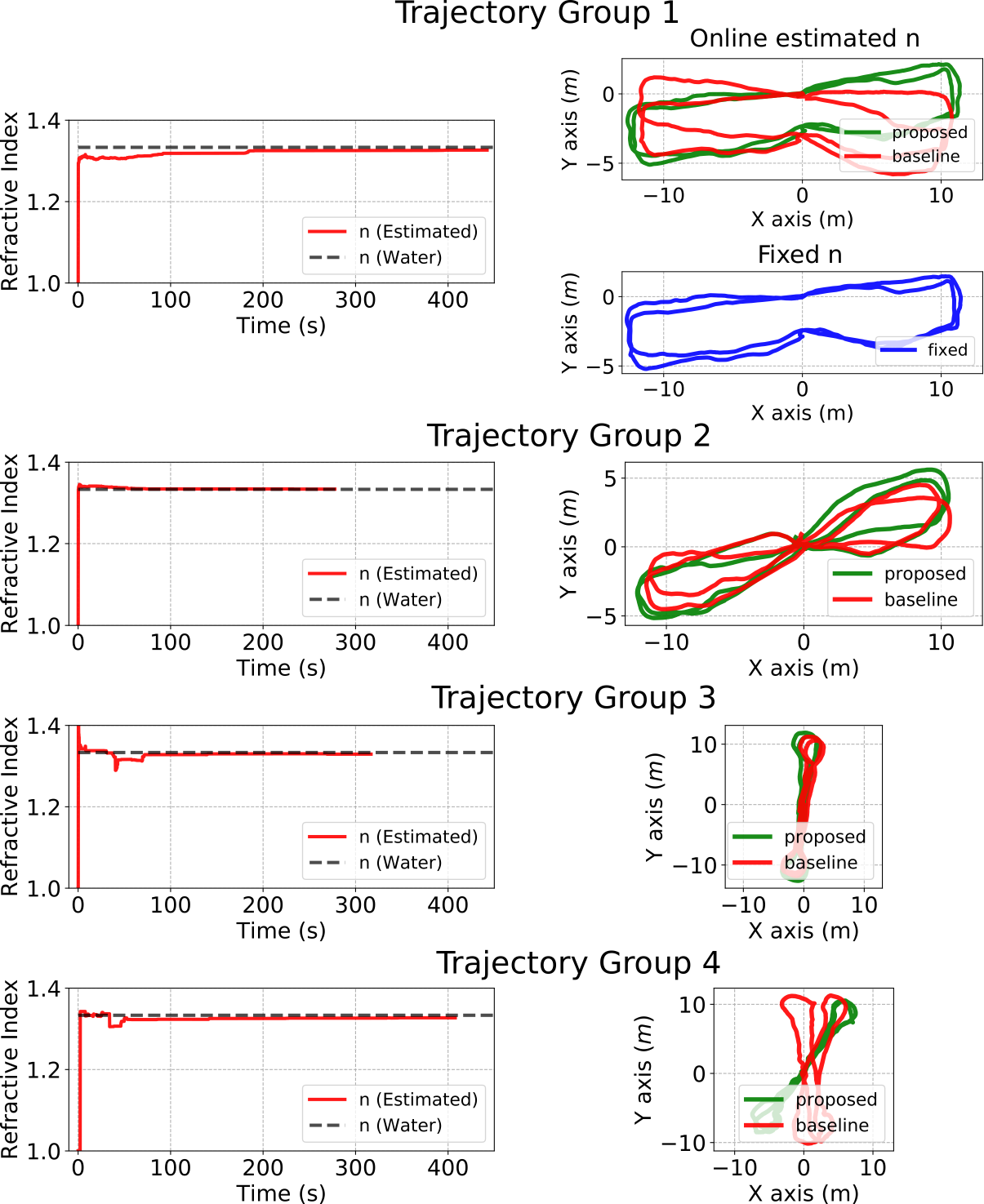}
\caption{Indicative refractive index $n$ estimation and odometry results from one of the trajectories of each of the groups of experiments conducted in the MC-lab. In the first case, the odometry result using the proposed camera model and fixed a priori information for the $n$ is also given. As shown the proposed approach offers significantly better results than ROVIO with a conventional camera model, while being on par with results assuming knowledge of the refractive index. }
\label{fig:collective_figure}
\end{figure}

The complete set of $24$ experiments were processed in a manner analogous to the one detailed above. Figure~\ref{fig:collective_figure} presents the results of one trajectory from each of the outlined groups, while Table~\ref{tab:collective_results_table} provides the mean estimate of the refractive index for each of the trajectory groups and the mean metric error versus the tracked AprilTags for our solution using the self-calibrating \ac{rcm}, a fixed \ac{rcm} incorporating a priori knowledge of the refractive index $n$, and the baseline ROVIO solution with the conventional camera model. As presented our approach leads to accurate estimation of the refractive index and significantly more accurate odometry estimation.

\begin{table}
\centering
\caption{Collective refractive index \& odometry estimation results}
    \label{tab:collective_results_table}
\begin{tabular}{|r|r|rrr|}
\hline
\multicolumn{1}{|l|}{\textbf{Trajectory Group}} & \multicolumn{1}{l|}{$n$ \textbf{(Estimated)}}  & \multicolumn{3}{c|}{\textbf{Trajectory Error}}                                                    \\ \hline
\multicolumn{1}{|l|}{}                    & \multicolumn{1}{l|}{1.33 for water} & \multicolumn{1}{l|}{Online} & \multicolumn{1}{l|}{Fixed} & \multicolumn{1}{l|}{Baseline} \\ \hline
\textbf{1} & 1.3326 & \multicolumn{1}{r|}{0.2306} & \multicolumn{1}{r|}{0.2585}  & 0.7751 \\ \hline
\textbf{2} & 1.3305 & \multicolumn{1}{r|}{0.3798} & \multicolumn{1}{r|}{0.3880} & 0.5618  \\ \hline
\textbf{3} & 1.3320 & \multicolumn{1}{r|}{0.3834}    & \multicolumn{1}{r|}{0.3570} & 0.8248  \\ \hline
\textbf{4} & 1.3157  & \multicolumn{1}{r|}{0.3668}  & \multicolumn{1}{r|}{0.3781} & 2.9946   \\ \hline
\end{tabular}
\end{table}

\subsection{Adaptation to Altering Media}

\begin{figure}[ht!]
\centering
    \includegraphics[width=0.99\columnwidth]{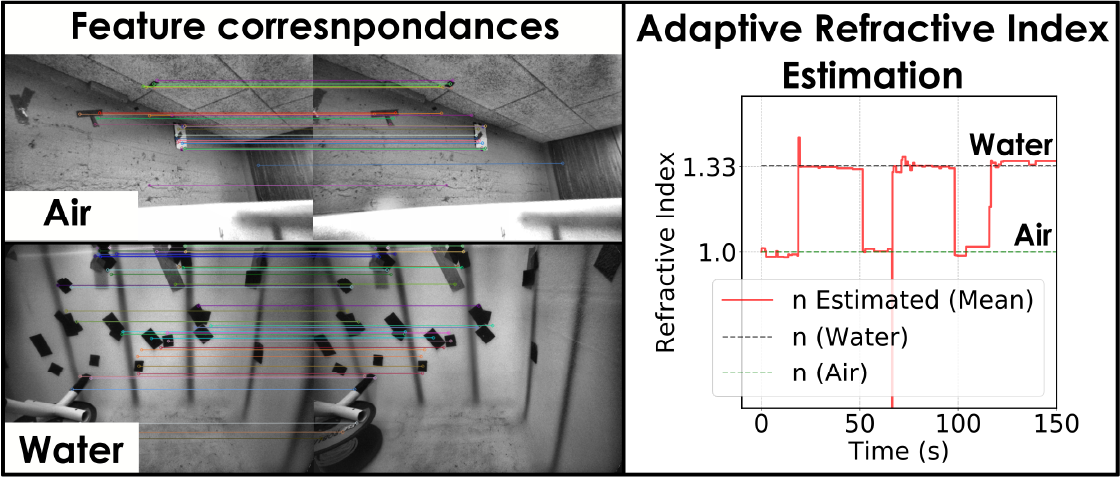}
\vspace{-2ex} 
\caption{Refractive index estimation while switching media (from air to water and back). The method presents robustness to this switching behavior.}
\label{fig:adaptive_refractive}
\end{figure}

For the case of non-smoothly varying media, the refractive index estimation process is adjusted to utilize a sliding window buffer $\mathbb{B}_{t,t-w_{s}}$ over a short horizon of length $w_{s}$ that allows to obtain a quick estimation of the refractive index when this is varying rapidly including in drastic changes such as the cameras exiting from the water. A relevant test was conducted where the camera system started outside the water, was moved inside and continued in such cycles. As shown in Figure~\ref{fig:adaptive_refractive} the method enabled the accurate tracking of the refractive index changes (from $n \approx 1.0$ to $n=1.33$ for air and water respectively) during this challenging experiment. Note that during medium switch, the past buffer of correspondences and $n$ is dropped and the buffer is then accumulating the estimation in the new media.

\section{Conclusions}\label{sec:concl}
This work presented a new refractive camera model and an approach for online estimation of the refractive index of a medium using stereo vision. The method was applied to the task of underwater visual-inertial odometry and demonstrated significant benefits compared to the common practice of laboriously calibrating cameras directly underwater. Using underwater visual-inertial data from a robot conducting $24$ trajectories, we demonstrate accurate estimation of refractive index and robust odometry. The method further demonstrates its performance in an experiment where the cameras are transitioning between air and water. The contribution allows underwater robots to be calibrated with less laborious processes and deployed reliably across different environments including in missions where due to change in pressure, temperature or other factors the refractive index is varying. 




\bibliographystyle{IEEEtran}
\bibliography{BIB/main,BIB/DepthPrediction,BIB/CamModel_and_MVG,BIB/GeneralUnderwater,BIB/VIO}

\begin{thebibliography}{10}
\providecommand{\url}[1]{#1}
\csname url@samestyle\endcsname
\providecommand{\newblock}{\relax}
\providecommand{\bibinfo}[2]{#2}
\providecommand{\BIBentrySTDinterwordspacing}{\spaceskip=0pt\relax}
\providecommand{\BIBentryALTinterwordstretchfactor}{4}
\providecommand{\BIBentryALTinterwordspacing}{\spaceskip=\fontdimen2\font plus
\BIBentryALTinterwordstretchfactor\fontdimen3\font minus
  \fontdimen4\font\relax}
\providecommand{\BIBforeignlanguage}[2]{{%
\expandafter\ifx\csname l@#1\endcsname\relax
\typeout{** WARNING: IEEEtran.bst: No hyphenation pattern has been}%
\typeout{** loaded for the language `#1'. Using the pattern for}%
\typeout{** the default language instead.}%
\else
\language=\csname l@#1\endcsname
\fi
#2}}
\providecommand{\BIBdecl}{\relax}
\BIBdecl

\bibitem{vasilijevic2017coordinated}
A.~Vasilijevi{\'c}, D.~Na{d}, F.~Mandi{\'c}, N.~Mi{\v{s}}kovi{\'c}, and
  Z.~Vuki{\'c}, ``Coordinated navigation of surface and underwater marine
  robotic vehicles for ocean sampling and environmental monitoring,''
  \emph{IEEE/ASME transactions on mechatronics}, vol.~22, no.~3, pp.
  1174--1184, 2017.

\bibitem{schill2018vertex}
F.~Schill, A.~Bahr, and A.~Martinoli, ``Vertex: A new distributed underwater
  robotic platform for environmental monitoring,'' in \emph{Distributed
  Autonomous Robotic Systems: The 13th International Symposium}.\hskip 1em plus
  0.5em minus 0.4em\relax Springer, 2018, pp. 679--693.

\bibitem{delmerico2019current}
J.~Delmerico, S.~Mintchev, A.~Giusti, B.~Gromov, K.~Melo, T.~Horvat, C.~Cadena,
  M.~Hutter, A.~Ijspeert, D.~Floreano \emph{et~al.}, ``The current state and
  future outlook of rescue robotics,'' \emph{Journal of Field Robotics},
  vol.~36, no.~7, pp. 1171--1191, 2019.

\bibitem{shukla2016application}
A.~Shukla and H.~Karki, ``Application of robotics in offshore oil and gas
  industry—a review part ii,'' \emph{Robotics and Autonomous Systems},
  vol.~75, pp. 508--524, 2016.

\bibitem{wu2019survey}
Y.~Wu, X.~Ta, R.~Xiao, Y.~Wei, D.~An, and D.~Li, ``Survey of underwater robot
  positioning navigation,'' \emph{Applied Ocean Research}, vol.~90, p. 101845,
  2019.

\bibitem{bahr2009cooperative}
A.~Bahr, J.~J. Leonard, and M.~F. Fallon, ``Cooperative localization for
  autonomous underwater vehicles,'' \emph{The International Journal of Robotics
  Research}, vol.~28, no.~6, pp. 714--728, 2009.

\bibitem{paull2013auv}
L.~Paull, S.~Saeedi, M.~Seto, and H.~Li, ``Auv navigation and localization: A
  review,'' \emph{IEEE Journal of oceanic engineering}, vol.~39, no.~1, pp.
  131--149, 2013.

\bibitem{xu2022robust}
Y.~Xu, R.~Zheng, S.~Zhang, and M.~Liu, ``Robust inertial-aided underwater
  localization based on imaging sonar keyframes,'' \emph{IEEE Transactions on
  Instrumentation and Measurement}, vol.~71, pp. 1--12, 2022.

\bibitem{johannsson2010imaging}
H.~Johannsson, M.~Kaess, B.~Englot, F.~Hover, and J.~Leonard, ``Imaging
  sonar-aided navigation for autonomous underwater harbor surveillance,'' in
  \emph{2010 IEEE/RSJ International Conference on Intelligent Robots and
  Systems}.\hskip 1em plus 0.5em minus 0.4em\relax IEEE, 2010, pp. 4396--4403.

\bibitem{ferrera2019aqualoc}
M.~Ferrera, V.~Creuze, J.~Moras, and P.~Trouv{\'e}-Peloux, ``Aqualoc: An
  underwater dataset for visual--inertial--pressure localization,'' \emph{The
  International Journal of Robotics Research}, vol.~38, no.~14, pp. 1549--1559,
  2019.

\bibitem{miao2021univio}
R.~Miao, J.~Qian, Y.~Song, R.~Ying, and P.~Liu, ``Univio: Unified direct and
  feature-based underwater stereo visual-inertial odometry,'' \emph{IEEE
  Transactions on Instrumentation and Measurement}, vol.~71, pp. 1--14, 2021.

\bibitem{teixeira2020deep}
B.~Teixeira, H.~Silva, A.~Matos, and E.~Silva, ``Deep learning for underwater
  visual odometry estimation,'' \emph{IEEE Access}, vol.~8, pp.
  44\,687--44\,701, 2020.

\bibitem{rahman2019svin2}
S.~Rahman, A.~Q. Li, and I.~Rekleitis, ``Svin2: An underwater slam system using
  sonar, visual, inertial, and depth sensor,'' in \emph{2019 IEEE/RSJ
  International Conference on Intelligent Robots and Systems (IROS)}.\hskip 1em
  plus 0.5em minus 0.4em\relax IEEE, 2019, pp. 1861--1868.

\bibitem{randall2023flsea}
Y.~Randall, ``Flsea: Underwater visual-inertial and stereo-vision
  forward-looking datasets,'' Ph.D. dissertation, University of Haifa (Israel),
  2023.

\bibitem{joshi2023sm}
B.~Joshi, H.~Damron, S.~Rahman, and I.~Rekleitis, ``Sm/vio: Robust underwater
  state estimation switching between model-based and visual inertial
  odometry,'' \emph{arXiv preprint arXiv:2304.01988}, 2023.

\bibitem{huang2017plate}
L.~Huang, X.~Zhao, S.~Cai, and Y.~Liu, ``Plate refractive camera model and its
  applications,'' \emph{Journal of Electronic Imaging}, vol.~26, no.~2, pp.
  023\,020--023\,020, 2017.

\bibitem{treibitz2011flat}
T.~Treibitz, Y.~Schechner, C.~Kunz, and H.~Singh, ``Flat refractive geometry,''
  \emph{IEEE transactions on pattern analysis and machine intelligence},
  vol.~34, no.~1, pp. 51--65, 2011.

\bibitem{sedlazeck2012perspective}
A.~Sedlazeck and R.~Koch, ``Perspective and non-perspective camera models in
  underwater imaging--overview and error analysis,'' in \emph{Outdoor and
  Large-Scale Real-World Scene Analysis: 15th International Workshop on
  Theoretical Foundations of Computer Vision, Dagstuhl Castle, Germany, June
  26-July 1, 2011. Revised Selected Papers}.\hskip 1em plus 0.5em minus
  0.4em\relax Springer, 2012, pp. 212--242.

\bibitem{haner2015absolute}
S.~Haner and K.~Astrom, ``Absolute pose for cameras under flat refractive
  interfaces,'' in \emph{Proceedings of the IEEE conference on computer vision
  and pattern recognition}, 2015, pp. 1428--1436.

\bibitem{hu2023refractive}
X.~Hu, F.~Lauze, and K.~S. Pedersen, ``Refractive pose refinement: Generalising
  the geometric relation between camera and refractive interface,''
  \emph{International Journal of Computer Vision}, vol. 131, no.~6, pp.
  1448--1476, 2023.

\bibitem{hu2021absolute}
X.~Hu, F.~Lauze, K.~S. Pedersen, and J.~M{\'e}lou, ``Absolute and relative pose
  estimation in refractive multi view,'' in \emph{Proceedings of the IEEE/CVF
  international conference on computer vision}, 2021, pp. 2569--2578.

\bibitem{gu2019environment}
C.~Gu, Y.~Cong, and G.~Sun, ``Environment driven underwater camera-imu
  calibration for monocular visual-inertial slam,'' in \emph{2019 International
  Conference on Robotics and Automation (ICRA)}.\hskip 1em plus 0.5em minus
  0.4em\relax IEEE, 2019, pp. 2405--2411.

\bibitem{zhang2021open}
P.~Zhang, Z.~Wu, J.~Wang, S.~Kong, M.~Tan, and J.~Yu, ``An open-source,
  fiducial-based, underwater stereo visual-inertial localization method with
  refraction correction,'' in \emph{2021 IEEE/RSJ International Conference on
  Intelligent Robots and Systems (IROS)}.\hskip 1em plus 0.5em minus
  0.4em\relax IEEE, 2021, pp. 4331--4336.

\bibitem{shkurti2011state}
F.~Shkurti, I.~Rekleitis, M.~Scaccia, and G.~Dudek, ``State estimation of an
  underwater robot using visual and inertial information,'' in \emph{2011
  IEEE/RSJ International Conference on Intelligent Robots and Systems}.\hskip
  1em plus 0.5em minus 0.4em\relax IEEE, 2011, pp. 5054--5060.

\bibitem{hu2022tightly}
C.~Hu, S.~Zhu, Y.~Liang, and W.~Song, ``Tightly-coupled
  visual-inertial-pressure fusion using forward and backward imu
  preintegration,'' \emph{IEEE Robotics and Automation Letters}, vol.~7, no.~3,
  pp. 6790--6797, 2022.

\bibitem{zuiderveld1994contrast}
K.~Zuiderveld, ``Contrast limited adaptive histogram equalization,''
  \emph{Graphics gems}, pp. 474--485, 1994.

\bibitem{lowe1999object}
D.~G. Lowe, ``Object recognition from local scale-invariant features,'' in
  \emph{Proceedings of the seventh IEEE international conference on computer
  vision}, vol.~2.\hskip 1em plus 0.5em minus 0.4em\relax Ieee, 1999, pp.
  1150--1157.

\bibitem{bloesch2015robust}
M.~Bloesch, S.~Omari, M.~Hutter, and R.~Siegwart, ``Robust visual inertial
  odometry using a direct ekf-based approach,'' in \emph{2015 IEEE/RSJ
  international conference on intelligent robots and systems (IROS)}.\hskip 1em
  plus 0.5em minus 0.4em\relax IEEE, 2015, pp. 298--304.

\bibitem{cerberus_science}
M.~Tranzatto, T.~Miki, M.~Dharmadhikari, L.~Bernreiter, M.~Kulkarni,
  F.~Mascarich, O.~Andersson, S.~Khattak, M.~Hutter, R.~Siegwart \emph{et~al.},
  ``Cerberus in the darpa subterranean challenge,'' \emph{Science Robotics},
  vol.~7, no.~66, p. eabp9742, 2022.

\bibitem{cerberus_finals}
M.~Tranzatto, M.~Dharmadhikari, L.~Bernreiter, M.~Camurri, S.~Khattak,
  F.~Mascarich, P.~Pfreundschuh, D.~Wisth, S.~Zimmermann, M.~Kulkarni
  \emph{et~al.}, ``Team cerberus wins the darpa subterranean challenge:
  Technical overview and lessons learned,'' \emph{arXiv preprint
  arXiv:2207.04914}, 2022.

\bibitem{cerberus_tunnel_urbam}
M.~Tranzatto, F.~Mascarich, L.~Bernreiter, C.~Godinho, M.~Camurri, S.~Khattak,
  T.~Dang, V.~Reijgwart, J.~Loeje, D.~Wisth \emph{et~al.}, ``Cerberus:
  Autonomous legged and aerial robotic exploration in the tunnel and urban
  circuits of the darpa subterranean challenge,'' \emph{arXiv preprint
  arXiv:2201.07067}, 2022.

\bibitem{lipson2021raftstereo}
L.~Lipson, Z.~Teed, and J.~Deng, ``Raft-stereo: Multilevel recurrent field
  transforms for stereo matching,'' in \emph{International Conference on 3D
  Vision (3DV)}, 2021.

\end{thebibliography}

\end{document}